# Graph-Based Blind Image Deblurring From a Single Photograph

Yuanchao Bai, *Student Member, IEEE,* Gene Cheung, *Senior Member, IEEE,* Xianming Liu, *Member, IEEE,*
Wen Gao, *Fellow, IEEE*

*Abstract*—**Blind image deblurring, *i.e.*, deblurring without knowledge of the blur kernel, is a highly ill-posed problem. The problem can be solved in two parts: i) estimate a blur kernel from the blurry image, and ii) given estimated blur kernel, deconvolve blurry input to restore the target image. In this paper, we propose a graph-based blind image deblurring algorithm by interpreting an image patch as a signal on a weighted graph. Specifically, we first argue that a skeleton image—a proxy that retains the strong gradients of the target but smooths out the details—can be used to accurately estimate the blur kernel and has a unique bi-modal edge weight distribution. Then, we design a reweighted graph total variation (RGTV) prior that can efficiently promote a bi-modal edge weight distribution given a blurry patch. Further, to analyze RGTV in the graph frequency domain, we introduce a new weight function to represent RGTV as a graph $l_1$-*Laplacian* regularizer. This leads to a graph spectral filtering interpretation of the prior with desirable properties, including robustness to noise and bias, strong piecewise smooth (PWS) filtering and sharpness promotion. Minimizing a blind image deblurring objective with RGTV results in a non-convex non-differentiable optimization problem. We leverage the new graph spectral interpretation for RGTV to design an efficient algorithm that solves for the skeleton image and the blur kernel alternately. Specifically for Gaussian blur, we propose a further speedup strategy for blind Gaussian deblurring using accelerated graph spectral filtering. Finally, with the computed blur kernel, recent non-blind image deblurring algorithms can be applied to restore the target image. Experimental results demonstrate that our algorithm successfully restores latent sharp images and outperforms state-of-the-art methods quantitatively and qualitatively.**

## I. INTRODUCTION

**I**Mage blur is a common image degradation, which is caused by out-of-focus photography or motions between objects and camera during exposure time. The blur process is usually modeled as

$$\mathbf{b} = \mathbf{x} \otimes \mathbf{k} + \mathbf{n}, \qquad (1)$$

where $\mathbf{b}$ is the observed blurry image, $\mathbf{x}$ is the latent sharp image, $\mathbf{k}$ is the blur kernel, $\mathbf{n}$ is the noise and $\otimes$ is the convolution operator. As an inverse problem, image deblurring is to recover the latent sharp image $\mathbf{x}$ from the blurry image

Yuanchao Bai and Wen Gao are with the School of Electronics Engineering and Computer Science, Peking University, Beijing, 100871, China. (e-mail: {yuanchao.bai, wgao}@pku.edu.cn).

Gene Cheung is with the National Institute of Informatics, Tokyo, 1018430, Japan (e-mail: cheung@nii.ac.jp).

Xianming Liu is with the School of Computer Science and Technology, Harbin Institute of Technology, Harbin, 150001, China (e-mail: csxm@hit.edu.cn).

$\mathbf{b}$. There are two categories of image deblurring depending on whether the blur kernel $\mathbf{k}$ is known, *i.e.*, non-blind image deblurring [1] and blind image deblurring [2], [3]. We focus on the blind image deblurring problem, where *both* the latent image $\mathbf{x}$ *and* the blur kernel $\mathbf{k}$ are unknown and must be restored given only the blurry image $\mathbf{b}$. It is a highly ill-posed problem, since the feasible solution of the problem is not only unstable but also non-unique.

To overcome the ill-posedness, for blind image deblurring, it is important to design a prior that promotes image sharpness and penalizes blurriness. However, conventional gradient-based priors of natural images tend to fail [3], because they usually favor blurry images with mostly low frequencies in the Fourier domain. Recently, many sophisticated image priors are proposed to deal with this problem, for example, $l_0$-norm based prior [4], low-rank prior [5] and dark channel prior [6]. Besides these priors, with the advance of *graph signal processing* (GSP) [7], graph-based priors have been designed for different image applications [8]–[11]. By modeling pixels as nodes with weighted edges that reflect inter-pixel similarities, images can be interpreted as signals on graphs. In this paper, we explore the relationship between graph and image blur, and propose a graph-based prior for blind image deblurring.

Specifically, instead of directly computing the natural image, we argue that a *skeleton image*—a piecewise smooth (PWS) proxy that retains the strong gradients of the target image but smooths out the details—is sufficient to estimate the blur kernel. We observe that, unlike blurry patches, the edge weights of a graph for the skeleton image have a unique bi-modal distribution. We thus propose a *reweighted graph total variation* (RGTV) prior to promote the desirable bi-modal distribution given a blurry patch. We juxtapose and analyze the advantages of RGTV against previous graph smoothness priors, such as graph total variation (GTV) [12]–[15] and the graph Laplacian regularizer [9] in the nodal domain. To analyze RGTV in graph frequency domain, we define a new graph weight function so that RGTV can be expressed as a graph $l_1$-*Laplacian regularizer*, similar in form to previous graph $l_2$-Laplacian regularizer [9]. Doing so means that RGTV can be interpreted as a low-pass graph spectral filter with desirable properties such as robustness to noise / blur and strong PWS filtering.

Based on our graph spectral interpretation of RGTV, we design an efficient algorithm that solves for the skeleton image and the blur kernel alternately. Moreover, specifically for Gaussian blur, we propose a further speedup strategy for blind Gaussian deblurring using accelerated graph spectral filtering [16]. Finally, with the estimated blur kernel $\mathbf{k}$, we de-convolve the blurry image using a non-blind deblurring method, like [4], [17], [18]. The contributions of this paper are summarized as follows:

1) We propose a graph-based image prior RGTV that promotes a bi-modal weight distribution to reconstruct a skeleton patch from a blurry observation, so that a suitable blur kernel can be simply derived thereafter.

2) We introduce a graph weight function so that GTV / RGTV can be expressed as a graph $l_1$-*Laplacian* regularizer. The prior can then be interpreted as a low-pass graph filter with desirable spectral properties. *To the best of our knowledge, we are the first to provide a graph frequency interpretation of GTV.*

3) Based on our spectral interpretation of RGTV, we design an efficient algorithm to solve the non-convex non-differentiable optimization problem alternately, where the two sub-problems to solve for the skeleton image and the blur kernel have closed-form solutions.

4) Specifically for Gaussian blur, we propose a further speedup strategy for blind Gaussian deblurring using accelerated graph spectral filters [16].

Experiments demonstrate that the proposed algorithm is competitive or even better than the state-of-the-art methods with lower complexity.

The outline of the paper is organized as follows. We review the related works on image deblurring and GSP in Sec. II. We introduce necessary GSP definitions and skeleton image in Sec. III. RGTV prior and its analysis in graph nodal and graph frequency domains are presented in Sec. IV. Blind deblurring algorithm is proposed in Sec. V. Experiments and conclusions are in Sec. VI and Sec. VII, respectively.

## II. RELATED WORK

### A. Image Deblurring

Depending on whether the blur kernel is known in advance or not, image deblurring is divided into two categories.

*1) Non-blind image deblurring:* Non-blind image deblurring [1] is to recover latent sharp image with known blur kernel, which has been studied for decades. It is an ill-posed problem since the inverse process is unstable to noise; even a small amount of noise will lead to severe distortions. Classic methods to solve this problem include Wiener filter [19] and Richardson-Lucy algorithm [20], [21]. In recent years, non-blind image deblurring is modeled as an optimization problem. Many famous image priors, such as total variation (TV) [22] and sparse priors [17], [23], have been introduced for regularization to deal with the ill-posedness in image restoration.

*2) Blind image deblurring:* Blind image deblurring [2], [3] is to recover a latent sharp image without knowledge of the underlying blur kernel. Conventional gradient-based priors of natural images tend to fail [3], because they usually favor blurry images with mostly low frequencies in the Fourier domain. With the progress of regularization and optimization, more sophisticated priors have been introduced to solve this problem, such as framelet based prior [24], $l_0$-norm based prior [4], [25], sparse-coding based prior [26], low-rank prior [5] and dark channel prior [6], *etc.* However, each has its own shortcomings. Framelet based prior [24] relies on manually crafted wavelet functions, which are not general enough to handle heterogeneous blurring scenarios. $l_0$-norm based prior [4], [25] is combinatorial in nature (thus non-convex), and its convex relaxation to $l_1$-norm requires sensitive parameter tuning for optimal performance. Sparse-coding based prior [26] assumes statistical similarity between the training set and the target image, which may not be true in practice. Low-rank prior [5] requires solving costly singular value decomposition (SVD), which has complexity $O(N^3)$ in general. Dark channel prior [6] is a comprehensive prior combining $l_0$-norm and non-linear dark channel computation, thus also suffers from high complexity.

Instead of the above mentioned priors, we investigate the potential of graph-based priors for the blind image deblurring problem. Our proposed graph-based prior RGTV is highly data-adaptive and can be efficiently solved using fast graph spectral filters via a novel interpretation of RGTV as a graph $l_1$-Laplacian regularizer.

### B. Graph Based Image Prior

GSP [7] is an emerging field to study signals on irregular data kernels described by graphs. By modeling pixels as nodes with weighted edges that reflect inter-pixel similarities, images (or image patches) can be interpreted as graph signals. For image restoration, graph based image priors, such as graph Laplacian regularizer [9] and GTV [12]–[15], have been designed for different inverse problems.

*1) Graph Laplacian Prior:* In [8], Hu *et al.* designed a scheme to soft-decode a JPEG-compressed PWS image by optimizing the desired graph signal and the similarity graph in a unified framework. In [10], Liu *et al.* soft-decoded JPEG-compressed natural images by using a combination of three priors, including a new graph smoothness prior called *Left Eigenvectors of Random walk Graph Laplacian* (LERaG), a compact dictionary trained by sparse representation and Laplacian distribution of discrete cosine transform coefficients. Pang *et al.* [9] analyzed graph Laplacian regularization in the continuous domain and provided insights for image denoising. With a doubly-stochastic graph Laplacian, Kheradmand *et al.* [11] developed a framwork for non-blind image deblurring.

*2) Graph Total Variation Prior:* GTV is one of the $p$-Dirichlet energy prior, which enjoys both desirable PWS-preserving properties and convexity. In [12], Elmoataz *et at.* analyzed the discrete $p$-Dirichlet energy in image and manifold processing. In [13], Hidane *et al.* employed GTV for non-linear multi-layered representation of graph signals. In [14], Couprie *et al.* proposed a dual constrained GTV regularization on graphs. In [15], Berger *et al.* used GTV to recover a smooth graph signal from noisy samples taken on a subset of graph nodes.

In this paper, we propose a novel RGTV prior to solve the blind image deblurring problem. The proposed method is a non-trivial extension of our recent work [27]: i) we provide analysis of GTV and RGTV in the graph spectral domain, the first to do so in the literature; ii) based on our spectral interpretation of RGTV, we design an efficient alternating iterative algorithm to solve the non-convex optimization problem; and iii) specifically for Gaussian blur, we propose a speedup strategy for blind Gaussian deblurring using accelerated graph spectral filtering [16].

## III. GSP DEFINITIONS AND SKELETON IMAGE

### A. Definitions in GSP

We first define GSP concepts needed in our work. A graph $\mathcal{G}(\mathcal{V}, \mathcal{E}, \mathbf{W})$ is a triple consisting of a finite set $\mathcal{V}$ of $N$ nodes (image pixels) and a finite set $\mathcal{E} \subset \mathcal{V} \times \mathcal{V}$ of $M$ edges. Each edge $(i, j) \in \mathcal{E}$ is undirected with a corresponding weight $w_{ij}$ which measures the similarity between nodes $i$ and $j$. Here we compute the weights using a Gaussian kernel [7]:

$$[\mathbf{W}]_{i,j} = w_{i,j} = \exp\left(-\frac{\|x_i - x_j\|^2}{\sigma^2}\right), \qquad (2)$$

where $\mathbf{W}$ is an *adjacency matrix* of size $N \times N$, $x_i$ and $x_j$ are the intensity values at pixels $i$ and $j$ of the image $\mathbf{x}$, and $\sigma$ is a parameter. $0 \leq w_{ij} \leq 1$ and the larger $w_{ij}$ is, the more similar the nodes $i$ and $j$ are to each other. Unlike the bilateral filter [28], we do not include Euclidean distance in the edge weight definition in (2), so that a denser graph would lead to stronger filtering among pixels of similar intensities. (Experimentally, graph density is chosen as a tradeoff between filtering strength and computation complexity.)

Given the adjacency matrix $\mathbf{W}$, a *combinatorial graph Laplacian matrix* $\mathbf{L}$ is a symmetric matrix defined as:

$$\mathbf{L} \triangleq \text{diag}(\mathbf{W1}) - \mathbf{W} \qquad (3)$$

where $\mathbf{1}$ is a vector of all 1's. $\text{diag}(\cdot)$ is an operator constructing a square diagonal matrix with the elements of input vector on the main diagonal.

Given (2) and (3), $\mathbf{L}$ is a real symmetric matrix, so there is an orthogonal matrix $\mathbf{U}$ that diagonalizes $\mathbf{L}$ via the Spectral Theorem,

$$\mathbf{L} = \mathbf{U}\mathbf{\Lambda}\mathbf{U}^T \qquad (4)$$

where $\mathbf{\Lambda}$ is a diagonal matrix containing eigenvalues $\lambda_k, k \in \{1, \ldots, N\}$. Each column $\mathbf{u}_k$ in $\mathbf{U}$ is an eigenvector corresponding to $\lambda_k$. Given $w_{i,j}$ is non-negative from (2), $\mathbf{L}$ is a positive semi-definite (PSD) matrix. Hence $\lambda_k \geq 0$ for each $k$ and $\mathbf{x}^T \mathbf{L} \mathbf{x} \geq 0$ for arbitrary graph signal $\mathbf{x}$. In the GSP literature [7], the non-negative eigenvalues $\lambda_k$ are interpreted as *graph frequencies* and corresponding eigenvectors in $\mathbf{U}$ as *graph frequency components*. Together, they define the *graph spectrum* for graph $\mathcal{G}$.

### B. Skeleton Image and its Bi-modal Weight Distribution

We define a *skeleton image*—a PWS version of the target image—as a proxy for the blind image deblurring problem. The skeleton image retains the strong gradients in a natural image but smooths out the minor details, which is similar to a structure extracted image [29] or an edge-aware smoothed image [30]. An illustrative example is shown in Fig. 1. Both the target natural image and its skeleton image are sharper than the blurry image in the middle.

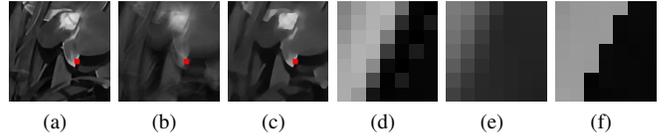

Fig. 1. Illustrations of different kinds of images. (a) a true natural image. (b) a blurry image. (c) a skeleton image. (d), (e) and (f) are patches in red squares of (a), (b) and (c), respectively.

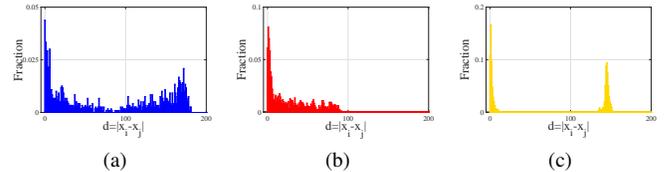

Fig. 2. Edge weight distribution around image edges. (a) a true natural patch. (b) a blurry patch. (c) a skeleton patch.

In order to differentiate between blurriness and sharpness in a pixel patch in a more mathematically rigorous manner, we further seek statistical descriptions of these patches. As an illustrative example, we construct a fully connected graph for each of three representative local patches (highlighted by red in Fig. 1) and examine its edge weight distribution, where edge weight $w_{i,j}$ is computed using (2). Fig. 2 shows the edge weight distributions (histograms) of the representative patches in Fig. 1d–1f. $x$-axis is the discrete inter-pixel difference $d = |x_i - x_j|$ for edge weight $w_{i,j}$; edge weight $w_{i,j}$ in (2) is a monotonically decreasing function of $d$. $y$-axis shows fractions of weights given $d$. We make the following key observation from the histograms:

> Both the target natural patch and its skeleton version have bi-modal distributions of edge weights, while the blurred patch does not, due to low-pass filtering during the blur process.

Bi-modal distribution means that the inter-pixel differences in an image patch are either very small or very large, *i.e.*, the patch is PWS. To be demonstrated experimentally later, the PWS skeleton patch is just as valuable as the target natural image in terms of computing an appropriate blur kernel, but the skeleton patch can be more easily reconstructed from a blurry patch than the natural patch.

Similar observations can be made on sparser graphs also. Given the desirable statistical property of the skeleton image patch, we next design a signal prior to *promote* a bi-modal distribution of edge weights given an observed blurry patch.

## IV. GRAPH-BASED IMAGE PRIOR AND ANALYSIS

### A. Reweighted Graph Total Variation Prior

We propose a *reweighted graph total variation* (RGTV) prior (regularizer) to promote the aforementioned bi-modal edge weight distribution in a target pixel patch. We first define the gradient operator of a graph signal $\mathbf{x}$. The gradient of node $i \in \mathcal{V}$ is defined as $\nabla_i \mathbf{x} \in \mathbb{R}^N$ and its $j$-th element is:

$$(\nabla_i \mathbf{x})_j \triangleq x_j - x_i, \tag{5}$$

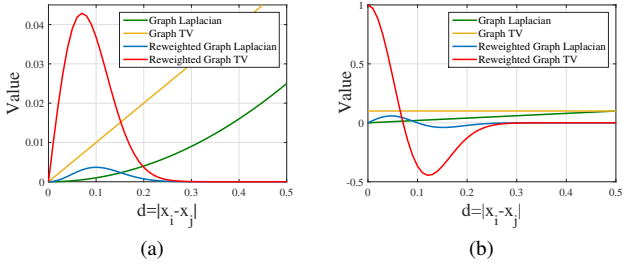

Fig. 3. Curves of regularizers and their corresponding first-derivatives for each $(i,j)$ pair. $d$ is normalized to $[0, 1]$. $w_{i,j} = 0.1$ for graph Laplacian and graph TV. $\sigma = 0.1$ for reweighted graph Laplacian and reweighted graph TV.

The *graph total variation* (GTV) [12]–[15] is defined as

$$\|\mathbf{x}\|_{GTV} = \sum_{i \in \mathcal{V}} \| \operatorname{diag}(\mathbf{W}_{i,\cdot}) \nabla_i \mathbf{x} \|_1$$
$$= \sum_{i=1}^{N} \sum_{j=1}^{N} w_{i,j} |x_j - x_i|, \tag{6}$$

where $\mathbf{W}_{i,\cdot}$ is the $i$-th row of the adjacency matrix $\mathbf{W}$. GTV initializes $\mathbf{W}$ using for example (2) and keeps it fixed, and hence does not promote bi-modal distribution of edge weights. Specifically, since (6) is separable, we can analyze the behavior of GTV using a single node pair $(i,j)$ separately like a two-node graph. With $d = |x_i - x_j|$ and fixed $w_{i,j}$, the regularizer for pair $(i,j)$ is $w_{i,j}d$, which is a linear function of $d$ with slope $w_{i,j}$. The curve of $w_{i,j}d$ has only one minimum at $d = 0$, as shown in Fig. 3a. Minimizing (6) only pushes $d$ towards 0, *i.e.*, smoothing the image $\mathbf{x}$.

Instead of using fixed $\mathbf{W}$, we extend the conventional graph TV to RGTV, where the graph weights $\mathbf{W}(\mathbf{x})$ are also functions of $\mathbf{x}$,

$$\|\mathbf{x}\|_{RGTV} = \sum_{i \in \mathcal{V}} \| \operatorname{diag}(\mathbf{W}_{i,\cdot}(\mathbf{x})) \nabla_i \mathbf{x} \|_1$$
$$= \sum_{i=1}^{N} \sum_{j=1}^{N} w_{i,j}(x_i, x_j) |x_j - x_i|, \tag{7}$$

where $\mathbf{W}_{i,\cdot}(\mathbf{x})$ is the $i$-th row of $\mathbf{W}(\mathbf{x})$ and $w_{i,j}(x_i, x_j)$ is the $(i,j)$ element of $\mathbf{W}(\mathbf{x})$. This extension has a fundamental difference, because the regularizer for pair $(i,j)$ now becomes $w_{i,j}(x_i, x_j)|x_j - x_i| = \exp(-d^2/\sigma^2) \cdot d$. The curve of this regularizer has one maximum at $\sigma/\sqrt{2}$ and two minima at 0 and $+\infty$, as shown in Fig. 3a. Minimizing (7) reduces $d$ if $d$ is smaller than $\sigma/\sqrt{2}$ and amplifies $d$ if $d$ is larger than $\sigma/\sqrt{2}$. *Thus, RGTV regularizer can effectively promote the desirable bi-modal edge weight distribution of sharp images.*

Using the aforementioned RGTV prior, we propose an optimization function for blind image deblurring in Sec. V.

### B. Comparisons with Graph Laplacian Prior

For comparison, we also analyze commonly used graph Laplacian regularizers. The graph Laplacian regularizer [9] is expressed as

$$\mathbf{x}^T \mathbf{L} \mathbf{x} = \sum_{i=1}^{N} \sum_{j=1}^{N} w_{i,j}(x_j - x_i)^2 \tag{8}$$

Like GTV, graph Laplacian initializes $\mathbf{L}$ and keeps it fixed, and hence does not promote the desirable bi-modal edge weight distribution. With $d = |x_i - x_j|$ and fixed $w_{i,j}$, the prior for each node pair $(i,j)$ is $w_{i,j}d^2$, which is a quadratic function of $d$ with coefficient $w_{i,j}$. The curve of $w_{i,j}d^2$ has only one minimum at $d = 0$, as shown in Fig. 3a. Minimizing (8) only pushes $d$ to 0, *i.e.*, smoothing the image $\mathbf{x}$.

We extend the conventional graph Laplacian to the reweighted graph Laplacian. Similar to RGTV, we define the reweighted graph Laplacian as

$$\mathbf{x}^T \mathbf{L}(\mathbf{x}) \mathbf{x} = \sum_{i=1}^{N} \sum_{j=1}^{N} w_{i,j}(x_i, x_j) \cdot (x_j - x_i)^2, \tag{9}$$

where Laplacian matrix $\mathbf{L}(\mathbf{x})$ is a function of $\mathbf{x}$ and $w_{i,j}(x_i, x_j)$ is the same as the definition in (7). Then, the regularizer for pair $(i,j)$ becomes $w_{i,j}(x_i, x_j)(x_j - x_i)^2 = \exp(-d^2/\sigma^2) \cdot d^2$. The curve has one maximum at $d = \sigma$ and two minima at 0 and $+\infty$, as shown in Fig. 3a. Thus the reweighted graph Laplacian *also* promotes the desirable bi-modal edge weight distribution, which explains its effectiveness in previous works on restoration of PWS images using this prior [8], [9], [31].

However, it has one significant drawback. Taking the first derivative of its function results in $\exp(-d^2/\sigma^2) \cdot 2d(1 - d^2/\sigma^2)$, as shown in Fig. 3b. $\lim_{d \to 0} \exp(-d^2/\sigma^2) \cdot 2d(1 - d^2/\sigma^2) = 0$, which means that the promotion of bi-modal edge weight distribution tends to slow down significantly when $d$ is close to 0 in practice.

Different from reweighted graph Laplacian, the first derivative of the cost function of RGTV is $\exp(-d^2/\sigma^2) \cdot (1 - 2d^2/\sigma^2)$, as shown in Fig. 3b. $\lim_{d \to 0} \exp(-d^2/\sigma^2) \cdot (1 - 2d^2/\sigma^2) = 1$, which means that RGTV can effectively promote a bi-modal edge weight distribution even as $d$ approaches 0. As an illustration, Fig. 4 compares the performance of reweighted graph Laplacian and the proposed RGTV in promoting bi-modal distribution from a blurry image experimentally. We observe that RGTV restores a better skeleton image with sharp edges than Reweighted graph Laplacian.

### C. Spectral Analysis of GTV and RGTV

As a signal smoothness prior defined with respect to the graph [7], GTV generalizes the well-known TV notion [22] and its edge-preserving property to the graph signal domain. However, the usage of $l_1$-norm means that there is no natural graph spectral interpretation—*i.e.*, promotion of certain low graph frequencies—like the graph Laplacian regularizer.

In this paper, we introduce a spectral interpretation of GTV by rewriting it as a novel $l_1$-*Laplacian operator* on a graph, inspired by [12]. Based on the spectral interpretation of GTV, we examine the graph spectrum of RGTV iteratively as the edge weights are updated across iterations. The desirable properties of RGTV become evident when carefully examining the low frequencies of its corresponding graph spectrum and its gradual transformation. Moreover, the new graph spectral interpretation of RGTV leads to an efficient algorithm for the non-convex and non-differentiable blind image deblurring problem, and an accelerated graph spectral filtering implementation specifically for Gaussian blur (to be discussed in Sec. V).

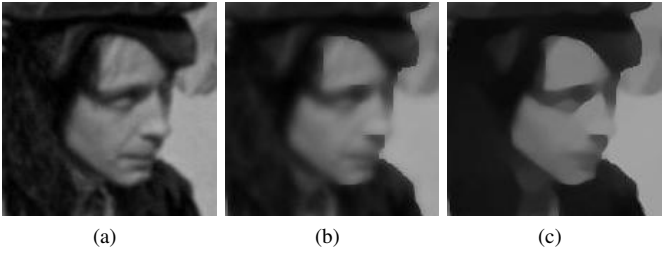

(a)      (b)      (c)

Fig. 4. (a) Image blurred by Gaussian blur, $\sigma_b = 1.5$. (b) Promoting bimodal distribution by reweighted graph Laplacian. (c) Promoting bi-modal distribution by RGTV. We set $\sigma = 0.1$ for reweighted graph Laplacian and $\sigma = 0.1 \times \sqrt{2}$ for RGTV to ensure the maximum of both priors are at $d = 0.1$.

Since a *graph spectrum* is defined with respect to a variation operator like the graph Laplacian, towards a spectral interpretation for GTV, we first define a new Laplacian operator for GTV. The conventional Laplacian operator on graph signal $\mathbf{x}$ can be deduced from the derivative of (8),

$$
\left( \partial \mathbf{x}^T \mathbf{L} \mathbf{x} \right)_i = 2 \cdot (\mathbf{L} \mathbf{x})_i = c \cdot \sum_{j=1}^{N} w_{i,j} \cdot (x_i - x_j),
$$
$$
= c \cdot \left( \sum_{j=1}^{N} w_{i,j} x_i - \sum_{j=1}^{N} w_{i,j} x_j \right) \quad (10)
$$

where $c$ is a coefficient derived from the derivative apart from the Laplacian operator.

Similarly, we take the *sub-derivative* of (6), as GTV is non-differentiable. By sub-differentiating and applying an upper-bound function to (6) near zero (detailed derivation is included in Appendix A), we get:

$$
\left( \partial \|\mathbf{x}\|_{GTV} \right)_i = c' \cdot \sum_{j=1}^{N} \gamma_{i,j} \cdot (x_i - x_j),
$$
$$
= c' \cdot \left( \sum_{j=1}^{N} \gamma_{i,j} x_i - \sum_{j=1}^{N} \gamma_{i,j} x_j \right) \quad (11)
$$

where $c'$ is a coefficient similar to $c$ in (10) and

$$
\gamma_{i,j} = \frac{w_{i,j}}{\max\{|x_j - x_i|, \epsilon\}} \quad (12)
$$

where $\epsilon$ is introduced as a small constant for numerical stability around 0. We see from the equation that when $|x_j - x_i| < \epsilon$, $\gamma_{i,j} = (1/\epsilon) w_{i,j}$, which is upper-bounded by $1/\epsilon$. In our experiments, we fix $\epsilon$ at 0.01.

Considering $\gamma_{i,j}$ as a new graph weight defined by (12), (11) is in the same form as (10). Hence we can define a new adjacency matrix $\mathbf{\Gamma}$ with the new weight function (12) and rewrite (11) in matrix form for GTV as

$$
\mathbf{L}_\Gamma \triangleq \operatorname{diag}(\mathbf{\Gamma} \mathbf{1}) - \mathbf{\Gamma}. \quad (13)
$$

We call $\mathbf{L}_\Gamma$ the $l_1$-*Laplacian* matrix. $\mathbf{L}_\Gamma$ is a real symmetric PSD matrix. With $\mathbf{L}_\Gamma$, we are able to analyze the spectrum of GTV like (4),

$$
\mathbf{L}_\Gamma = \mathbf{U}_\Gamma \mathbf{\Lambda}_\Gamma \mathbf{U}_\Gamma^T \quad (14)
$$

where $\mathbf{\Lambda}_\Gamma$ is a diagonal matrix containing graph frequencies of GTV, and $\mathbf{U}_\Gamma$ contains corresponding graph frequency components as columns. $\min \{\operatorname{diag}(\mathbf{\Lambda}_\Gamma)\}$ is 0 and $\max \{\operatorname{diag}(\mathbf{\Lambda}_\Gamma)\}$ is upper-bounded by $\max_i \left( \frac{2 d_i}{\epsilon} \right)$, where $d_i$ is the degree of node $i$, as proven in Appendix B. The bounds of $\operatorname{diag}(\mathbf{\Lambda}_\Gamma)$ are used in later stability analysis of our algorithm in Sec. V.

For clarity of presentation, in the following we denote by $\mathbf{L}_W$ the conventional graph Laplacian matrix computed from adjacency matrix $\mathbf{W}$, and denote by $\mathbf{L}_\Gamma$ the $l_1$-*Laplacian* matrix computed from adjacency matrix $\mathbf{\Gamma}$.

Fig. 5 and Fig. 6 show experiment results for a simple one-dimensional PWS signal, demonstrating that the spectrum of GTV has better PWS properties than conventional graph Laplacian. First, the PWS property of the second eigenvector (lowest AC frequency component) of GTV is more robust than the graph Laplacian regularizer when noise and/or blur are applied to the ideal signal, as shown in Fig. 5d, 5e and 5f.

Second, we claim that GTV is a stronger PWS-preserving filter than the graph Laplacian regularizer; we show this by examining the *relative eigenvalues* $\lambda_k / \lambda_2$. Specifically, consider the known graph-spectral filter from a standard MAP formulation [9], where the nodal domain filter $\mathbf{x}^* = (\mathbf{I} + \mu \cdot \mathbf{L}_{\{W, \Gamma\}})^{-1} \mathbf{y}$ can be expressed in the graph frequency domain as:

$$
\mathbf{x}^* = \mathbf{U}_{\{W, \Gamma\}} \operatorname{diag} \left( \frac{1}{1 + \mu \cdot \lambda_k^{\{W, \Gamma\}}} \right) \mathbf{U}_{\{W, \Gamma\}}^T \mathbf{y}, \quad (15)
$$

where $\{W, \Gamma\}$ means either the conventional graph Laplacian operator or our proposed $l_1$-*Laplacian* operator is employed. $\mu$ is a parameter trading off the importance of the fidelity term and the graph-signal smoothness prior in the original MAP formulation, and $k \in \{1, \ldots, N\}$. In Fig. 6, we observe that $\lambda_1^{\{W, \Gamma\}} / \lambda_2^{\{W, \Gamma\}} = 0$ and $\lambda_2^{\{W, \Gamma\}} / \lambda_2^{\{W, \Gamma\}} = 1$ for both GTV and graph Laplacian, but $\lambda_k^\Gamma / \lambda_2^\Gamma$ of GTV for $k > 2$ is much larger than $\lambda_k^W / \lambda_2^W$ of the graph Laplacian regularizer, meaning that GTV penalizes high graph frequencies more severely than the graph Laplacian regularizer.

**Spectral Analysis of RGTV:** Based on the spectral analysis of GTV above, we further analyze the behaviour of the spectrum of RGTV. RGTV inherits the desirable spectral properties—robustness to noise and blur, and strong PWS filtering—from GTV *and* promotes bi-modal edge weight distribution at the same time. Since the adjacency matrix $\mathbf{W}(\mathbf{x})$ of RGTV is a function of signal $\mathbf{x}$, there is no fixed spectrum for RGTV. Instead, we first initialize weights for RGTV and compute its spectrum like GTV, and then we update the weights and examine the gradual transformation of spectrum iteratively. Considering the graph-spectral filter again, the ideal RGTV filter satisfies $(\mathbf{I} + \mu \cdot \mathbf{L}_\Gamma(\mathbf{x}^*)) \, \mathbf{x}^* = \mathbf{y}$. We initialize $\mathbf{x}_{(0)} = \mathbf{y}$ and perform the following filter (16) iteratively until convergence,

$$\mathbf{x}^{(n+1)} = \mathbf{U}_\Gamma^{(n)} \operatorname{diag}\left(\frac{1}{1 + \mu \cdot \lambda_k^\Gamma(\mathbf{x}^{(n)})}\right) \mathbf{U}_\Gamma^{(n)^T} \mathbf{y} \quad (16)$$

where $\mathbf{U}_\Gamma^{(n)} = \mathbf{U}_\Gamma(\mathbf{x}^{(n)})$.

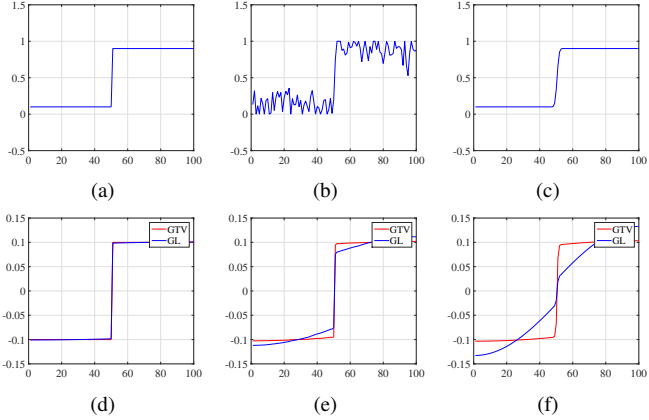

Fig. 5. Illustrative experiments of graph spectrum on 1D graph signals. (a) an ideal PWS signal. (b) a PWS signal with Gaussian noise, $\sigma_n = 0.02$. (c) a PWS signal blurred by a Gaussian blur, $\sigma_b = 1$. (d), (e) and (f) are the lowest AC frequency components of GTV and GL (graph Laplacian) corresponding to (a), (b) and (c). The graphs are constructed as a 4-neighbour adjacency matrix with weight parameter $\sigma = 0.3$.

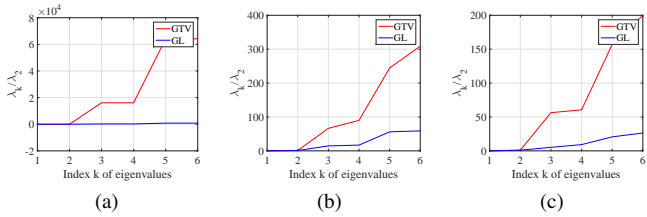

Fig. 6. Relative eigenvalues. (a), (b) and (c) represent the curves of relative eigenvalue $\lambda_k / \lambda_2$ of GTV and GL (graph Laplacian) with respect to $k$. Each plot corresponds to the signal in Fig. 5a, Fig. 5b and Fig. 5c, respectively.

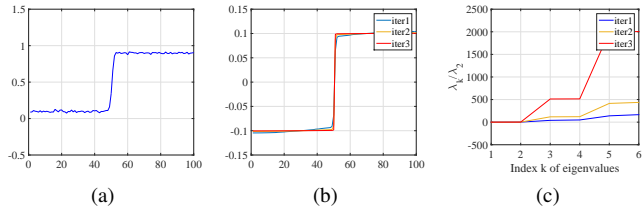

Fig. 7. Illustrative experiments of graph spectrum of RGTV on a 1D graph signal. (a) a PWS signal blurred by a Gaussian blur $\sigma_b = 1$ with Gaussian noise $\sigma_n = 0.0001$. (b) is the second eigenvectors of RGTV in each iteration. (c) represents the curves of relative eigenvalue $\lambda_k / \lambda_2$ with respect to $k$ in each iteration. The graphs are constructed as a 4-neighbour adjacency matrix with weight parameter $\sigma = 0.3$.

A one-dimensional illustrative experiment is shown in Fig. 7. The first AC graph frequency component of RGTV becomes sharper (Fig. 7b), and the low-pass graph filtering strength becomes larger (Fig. 7c) with iteration, which demonstrates the edge-sharpness promotion property of RGTV from the graph frequency perspective.

From both the analysis on the graph nodal and graph spectral domain, we can theoretically conclude that RGTV is an effective prior for the blind image deblurring, and more effective than previous graph smoothness priors, such as graph Laplacian regularizer [9]–[11] and conventional GTV [13]–[15].

## V. BLIND IMAGE DEBLURRING ALGORITHM

Using the blur image model in (1), we pose the blind image deblurring problem as an optimization as follows using our proposed RGTV prior:

$$\hat{\mathbf{x}}, \hat{\mathbf{k}} = \arg\min_{\mathbf{x}, \mathbf{k}} \frac{1}{2} \|\mathbf{x} \otimes \mathbf{k} - \mathbf{b}\|_2^2 + \beta \|\mathbf{x}\|_{RGTV} + \mu \|\mathbf{k}\|_2^2 \quad (17)$$

where the first term is the data fidelity term, and the remaining two terms are regularization terms for variables $\mathbf{x}$ and $\mathbf{k}$, respectively. $\beta$ and $\mu$ are two corresponding parameters.

The optimization (17) is non-convex and non-differentiable, which is challenging to solve. Here we apply a *coarse-to-fine strategy* [32] to solve (17), in order to make the solver robust even for large blur kernels. In the coarse-to-fine strategy, we construct an image pyramid by down-sampling the blurry image and perform blind image deblurring scale-by-scale. In each scale, we estimate $\hat{\mathbf{k}}$ and $\hat{\mathbf{x}}$ alternatively, and then up-sample $\hat{\mathbf{k}}$ as the initial value for the finer scale. The optimization algorithm in each scale is sketched in Algorithm 1.

---

**Algorithm 1** Blind Deblurring Algorithm in Each Scale

**Input:** Blurry image $\mathbf{b}$ and kernel size $h \times h$.
1: Initialize $\hat{\mathbf{k}}$ with delta function or the result from coarser scale.
2: **while** not converge **do**
  Compute $\hat{\mathbf{x}}$ by solving (18).
  Compute $\hat{\mathbf{k}}$ by solving (21).
  $\beta \leftarrow \beta / 1.1$.
 **endwhile**
**Output:** Estimated blur kernel $\hat{\mathbf{k}}$ and skeleton image $\hat{\mathbf{x}}$.

---

The minimizer $\hat{\mathbf{x}}$ is our PWS proxy—the skeleton image—in order to estimate a good blur kernel $\hat{\mathbf{k}}$. To restore the natural sharp image given estimated blur kernel $\hat{\mathbf{k}}$, we can use recent non-blind image deblurring algorithms to deblur the blurry image $\mathbf{b}$ such as [4], [17], [18].

### A. Skeleton Image Restoration

Given $\hat{\mathbf{k}}$, optimization (17) to solve $\mathbf{x}$ becomes:

$$\hat{\mathbf{x}} = \arg\min_{\mathbf{x}} \frac{1}{2} \|\mathbf{x} \otimes \hat{\mathbf{k}} - \mathbf{b}\|_2^2 + \beta \|\mathbf{x}\|_{RGTV} \quad (18)$$

RGTV is a non-differentiable and non-convex prior, where the edge weights are functions of $\mathbf{x}$. To solve (18), we leverage on the spectral analysis in Sec. IV-C and employ an alternating scheme with the proposed $l_1$-*Laplacian* of GTV to approximate RGTV; *i.e.*, we first optimize $\mathbf{x}$ with initialized $\mathbf{L}_\Gamma$, then we update $\mathbf{L}_\Gamma$ with $\mathbf{L}_\Gamma(\hat{\mathbf{x}})$ and optimize $\mathbf{x}$ again. The alternating algorithm runs iteratively until convergence as the solution to (18). The steps of solving (18) is summarized in Algorithm 2.

Fixing $\mathbf{L}_\Gamma$ and $\hat{\mathbf{k}}$ to solve for $\mathbf{x}$, the problem becomes a non-blind image deblurring problem with a graph Laplacian regularizer:

$$\hat{\mathbf{x}} = \arg\min_{\mathbf{x}} \frac{1}{2}\|\mathbf{x} \otimes \hat{\mathbf{k}} - \mathbf{b}\|_2^2 + \beta \cdot \mathbf{x}^T \mathbf{L}_\Gamma \mathbf{x} \quad (19)$$

As (19) is a quadratic convex optimization function, it is equivalent to solving the following system of linear equations,

$$(\hat{\mathbf{K}}^T\hat{\mathbf{K}} + 2\beta \cdot \mathbf{L}_\Gamma)\hat{\mathbf{x}} = \hat{\mathbf{K}}^T\mathbf{b} \quad (20)$$

where $\hat{\mathbf{K}}$ is a block circulant with circulant blocks (BCCB) matrix that is the matrix representation of convolving with $\hat{\mathbf{k}}$. The matrix $\hat{\mathbf{K}}^T\hat{\mathbf{K}} + 2\beta \cdot \mathbf{L}_\Gamma$ is a real symmetric positive-definite matrix, proven in Appendix C. Further, one can verify if (20) is well-conditioned numerically via a *Power Method* [33] to check its condition number[1]: running Power Method twice to compute the maximum and minimum eigenvalues of $\hat{\mathbf{K}}^T\hat{\mathbf{K}}+2\beta\cdot\mathbf{L}_\Gamma$ and checking the condition number $\lambda_{\max}/\lambda_{\min}$. We found that the matrix typically has small condition number in our experiments. In the rare case when the condition number is large, for stability we can add an *iterative refinement term* $\epsilon\,\mathbf{I}$ and solve (20) iteratively, details described in pg.146 [34].

Since the left-hand-side matrix is sparse, positive definite and symmetric, we can solve (20) efficiently using the *Conjugate Gradient* (CG) method [35]. In practice, we can implement the $\hat{\mathbf{K}}\mathbf{x}$ as 2D convolution and accelerate it with *Fast Fourier Transform* (FFT), and implement $\mathbf{L}_\Gamma\mathbf{x}$ as locally graph filter, instead of matrix computation.

---

**Algorithm 2** Solving (18)

---

**Input:** Blurry image $\mathbf{b}$ and estimated kernel $\hat{\mathbf{k}}$.
1: Initialize $\mathbf{L}_\Gamma$ as an unweighted graph Laplacian.
2: **while** not converge **do**
    Update $\hat{\mathbf{x}}$ by solving (20) with CG.
    Update $\mathbf{L}_\Gamma = \mathbf{L}_\Gamma(\hat{\mathbf{x}})$ using (12) and (13).
  **endwhile**
**Output:** Restored skeleton image $\hat{\mathbf{x}}$.

---

### B. Blur Kernel Estimation

To solve $\mathbf{k}$ given $\hat{\mathbf{x}}$, we make a slight modification by solving $\mathbf{k}$ in the gradient domain to avoid artifacts [36], [37]. The optimization (17) becomes:

$$\hat{\mathbf{k}} = \arg\min_{\mathbf{k}} \frac{1}{2}\|\nabla\hat{\mathbf{x}} \otimes \mathbf{k} - \nabla\mathbf{b}\|_2^2 + \mu\|\mathbf{k}\|_2^2 \quad (21)$$

where $\nabla$ is the gradient operator. (21) is a quadratic convex function and has a closed-form solver like deconvolution. We accelerate the solver via FFT [36]. After obtaining $\hat{\mathbf{k}}$, we threshold the negative elements to zeros and normalize $\hat{\mathbf{k}}$ to ensure $\sum_i \hat{k}_i = 1$.

The rationale for successful kernel estimation with skeleton image $\hat{\mathbf{x}}$ is that (21) is an over-determined function. Because the kernel $\mathbf{k}$ is much smaller than the image $\mathbf{x}$, the skeleton image $\hat{\mathbf{x}}$ with restored sharp edges is enough for kernel estimation.

---
[1] https://en.wikipedia.org/wiki/Condition_number#Matrices

### C. Acceleration for Specific Gaussian Blur Deblurring

Gaussian blur is a widely-assumed blur type in image restoration applications, such as out-of-focus deblurring or image super-resolution [38]–[40]. Assuming we know *a priori* that the blur type is Gaussian or close to Gaussian, we propose an acceleration for blind Gaussian blur deblurring.

In Algorithm 1, solving (18) with general blur kernel $\hat{\mathbf{k}}$ takes most of the running time. Under the assumption of Gaussian blur, we replace $\hat{\mathbf{k}}$ (or $\hat{\mathbf{K}}$) with graph filter $\mathbf{I} + a \cdot \mathbf{L}_\Gamma$, where $\mathbf{L}_\Gamma$ is first initialized as an unweighted graph Laplacian. The filter $\mathbf{I} + a \cdot \mathbf{L}_\Gamma$ with $a < 0$ is a smoothing process, which can be considered as an approximation of Gaussian blur. To set a suitable initial value for parameter $a$, we manually blur several sharp images with Gaussian blurs $\sigma_b \in (0, 2]$, as $\sigma_b \in (0, 2]$ are commonly used in practice. We learn the optimal $a = -0.07$ from the sharp and blurred image pairs using the *least square method*,

$$a = \arg\min_{a} \|(\mathbf{I} + a \cdot \mathbf{L}_\Gamma)\mathbf{X} - \mathbf{Y}\|_2^2. \quad (22)$$

where matrix $\mathbf{X} = [\mathbf{x}_1, \mathbf{x}_2, ..., \mathbf{x}_n]$ represents $n$ sharp images, matrix $\mathbf{Y} = [\mathbf{y}_1, \mathbf{y}_2, ..., \mathbf{y}_n]$ represents corresponding blurred images.

With $\mathbf{I} + a \cdot \mathbf{L}_\Gamma$, the skeleton image restoration function (18) is modified to (23),

$$\hat{\mathbf{x}} = \arg\min_{\mathbf{x}} \frac{1}{2}\|(\mathbf{I} + a \cdot \mathbf{L}_\Gamma)\mathbf{x} - \mathbf{b}\|_2^2 + \beta\|\mathbf{x}\|_{RGTV}. \quad (23)$$

The advantage of (23) is that $\mathbf{I} + a \cdot \mathbf{L}_\Gamma$ and graph Laplacian $\mathbf{x}^T\mathbf{L}_\Gamma\mathbf{x}$ in (19) now share the same graph frequency bases. The closed-form solution (20) now becomes:

$$\begin{aligned}
\hat{\mathbf{x}} &= \left(\frac{g(\mathbf{L}_\Gamma)}{g^2(\mathbf{L}_\Gamma) + 2\beta \cdot \mathbf{L}_\Gamma}\right)\mathbf{b} \\
&= \mathbf{U}_\Gamma\left(\frac{g(\mathbf{\Lambda}_\Gamma)}{g^2(\mathbf{\Lambda}_\Gamma) + 2\beta \cdot \mathbf{\Lambda}_\Gamma}\right)\mathbf{U}_\Gamma^T\mathbf{b}
\end{aligned} \quad (24)$$

where $g(\mathbf{X}) = \mathbf{I} + a \cdot \mathbf{X}$. (24) is a polynomial graph filter to signal $\mathbf{b}$ and can be implemented with an accelerated *Lanczos method* [16], which is faster than solving (20) with CG for this specific problem. *Lanczos method* computes an orthonormal basis $\mathbf{V}_Z = [v_1, ..., v_Z]$ of the Krylov subspace $\mathbf{K}_Z(\mathbf{L}_\Gamma, \mathbf{b}) = \text{span}\{\mathbf{b}, \mathbf{L}_\Gamma\mathbf{b}, ..., \mathbf{L}_\Gamma^{Z-1}\mathbf{b}\}$ and the corresponding symmetric scalar tridiagonal matrix $\mathbf{H}_Z$:

$$\mathbf{V}_Z^* \, \mathbf{L}_\Gamma \mathbf{V}_Z = \mathbf{H}_Z = \begin{bmatrix} \alpha_1 & \beta_2 & & & \\ \beta_2 & \alpha_2 & \beta_3 & & \\ & \beta_3 & \alpha_3 & \ddots & \\ & & \ddots & \ddots & \beta_M \\ & & & \beta_M & \alpha_M \end{bmatrix} \quad (25)$$

The approximation of $\hat{\mathbf{x}}$ with order $Z$ *Lanczos method* is:

$$\hat{\mathbf{x}} = f(\mathbf{L}_\Gamma)\mathbf{b} \approx \|\mathbf{b}\|_2\mathbf{V}_Z f(\mathbf{H}_Z)e_1 := f_Z, \quad (26)$$

where $e_1 \in \mathbb{R}_Z$ is the first unit vector. $f(\mathbf{H}_Z)$ is inexpensive given $Z \ll N$.

We update $\hat{\mathbf{x}}$, $\mathbf{L}_\Gamma = \mathbf{L}_\Gamma(\hat{\mathbf{x}})$ and parameter $a$ using (27) iteratively until convergence,

$$a = \arg\min_{a} \|(\mathbf{I} + a \cdot \mathbf{L}_\Gamma)\hat{\mathbf{x}} - \mathbf{b}\|_2^2. \quad (27)$$

---

**Algorithm 3** Accelerated Blind Gaussian Blur Deblurring

---

**Input:** Blurry image $\mathbf{b}$ and kernel size $h \times h$.
1: Initialize $\mathbf{L}_\Gamma$ as an unweighted graph Laplacian.
   Initialize blur with $\mathbf{I} + a \cdot \mathbf{L}_\Gamma$ smoothing.
2: Computing $\hat{\mathbf{x}}$ by solving (23):
   **while** not converge **do**
      Update $\hat{\mathbf{x}}$ using Lanczos method (25) and (26).
      Update $\mathbf{L}_\Gamma = \mathbf{L}_\Gamma(\hat{\mathbf{x}})$ using (12) and (13).
      Update $a$ using (27).
   **endwhile**
3: Compute $\hat{\mathbf{k}}$ by solving (21).
**Output:** Estimated blur kernel $\hat{\mathbf{k}}$ and skeleton image $\hat{\mathbf{x}}$.

---

Afterwards, a satisfactory skeleton image $\hat{\mathbf{x}}$ can be restored and then we can compute the blur kernel $\hat{\mathbf{k}}$ using (21), as shown in Algorithm 3.

## VI. EXPERIMENTS AND DISCUSSIONS

In this section, we set up comprehensive experiments to verify the effectiveness of our proposed algorithms in solving various blind image deblurring problems. The proposed algorithms are evaluated on three kinds of blurred cases, including artificial blurred database, real motion blurred images, and Gaussian blurred images. For all cases, we compare the performance of the proposed algorithms with the best existing blind image deblurring algorithms in the literature. We first evaluate the performance of Algorithm 1 on the artificial blurred database [26], which is a widely used database appropriate for both qualitative and quantitative assessment. Further, we apply Algorithm 1 on real motion blurred images to demonstrate its practicality. In the third part, we consider specific Gaussian blur introduced by manual blurring or image scaling. We evaluate the accelerated Algorithm 3 in this part. All the experiments are implemented on the Matlab 2015a platform with i7-4765T CPU.

We tune the parameters of the proposed Algorithm 1 based on the artificial blurred database [26] and find them satisfactory for almost all the cases. The down-sampling factor for coarse-to-fine strategy is set to $\log_2 3$. We construct a four-neighbour adjacency graph on the image as a trade-off between performance and computation complexity. In (2), $\sigma = 0.1$. In (17), $\beta = 0.01$ and $\mu = 0.05$. The same parameters are used for Algorithm 3. In blind image deblurring, the *kernel size* is unknown and is an important parameter. For fair comparisons, we set the same kernel size for all the algorithms to estimate the blur kernel in each case. Then, we use the same non-blind image deblurring algorithm to reconstruct sharp images with estimated blur kernels. The detail settings are described in each experiment.

### A. Artificial Blurred Database

We evaluate the qualitative and quantitative performance of the proposed algorithm on a large database introduced by Sun *et al.* [26]. The database consists of 640 gray-scale blurry images (typically $1024 \times 768$), which were made by convolving 80 high-resolution sharp images with 8 blur kernels from Levin *et al.* [3] and were added $1\%$ white Gaussian noise.

TABLE I
QUANTITATIVE COMPARISONS OF ALL METHODS OVER THE ENTIRE DATABASE.

| Blind Deblurring Method | Mean error ratio | Worst error ratio | Success rate ($r \le 5$) |
|---|---|---|---|
| Cho *et al.* [43] | 28.1 | 165.0 | 11.7% |
| Krishnan *et al.* [42] | 11.7 | 133.2 | 24.8% |
| Levin *et al.* [41] | 6.6 | 40.9 | 46.7% |
| Cho & Lee [36] | 8.7 | 111.1 | 65.5% |
| Xu & Jia [37] | 3.6 | 65.3 | 85.8% |
| Sun *et al.* [26] | 2.5 | 30.5 | 93.4% |
| Michaeli & Irani [44] | 2.6 | 9.3 | 95.9% |
| Lai *et al.* [45] | 2.1 | 17.9 | 97.3% |
| Ours + [18] | **2.0** | **9.2** | **99.7%** |
| Pan *et al.* [6] | 1.6 | 8.8 | 99.1% |
| Ours + [4] | **1.5** | **5.1** | **99.8%** |

We compare with nine best existing blind image deblurring algorithms, *i.e.*, Levin *et al.* [41], Krishnan *et al.* [42], Cho *et al.* [43], Cho & Lee [36], Xu & Jia [37], Sun *et al.* [26], Michaeli & Irani [44], Lai *et al.* [45] and Pan *et al.* [6]. The results of the first eight algorithms, except Pan *et al.* [6], are offered by Sun *et al.* [26] or their authors on their websites. These algorithms estimate blur kernels and then use non-blind deblurring algorithm [18] to restore the latent sharp images as the final step. In the experiments, *kernel size* is set to $51 \times 51$ by all these algorithms. The code of Pan *et al.* [6] is provided online, so we run their code on the database. To fairly compare with Pan *et al.* [6], we do not modify their codes but use the same setting as theirs to run the proposed algorithm again, *i.e.*, *kernel size* $= 51 \times 51$ and use sophisticated [4] as the final non-blind deblurring algorithm.

Considering the upper-bound performance of blind image deblurring is non-blind image deblurring with ground-truth blur kernel, we measure the quantitative performances of all the algorithms with *error ratio*, introduced by Levin *et al.* [3],

$$r = \frac{\|\mathbf{x} - \mathbf{x}_{\hat{\mathbf{k}}}\|^2}{\|\mathbf{x} - \mathbf{x}_{\mathbf{k}}\|^2}, \quad (28)$$

where $\mathbf{x}_{\hat{\mathbf{k}}}$ and $\mathbf{x}_{\mathbf{k}}$ are the images restored by estimated kernel $\hat{\mathbf{k}}$ and ground-truth kernel $\mathbf{k}$, respectively. $\mathbf{x}$ is the ground-truth sharp image. $r \ge 1$ and the smaller, the better. Same as Michaeli & Irani [44], we assume that $r \le 5$ is the threshold to decide the success of deblurred results.

Fig. 8 reports the fractions of images that can be restored under different error ratios. Our algorithm and [6], [26], [44], [45] are close when $r \le 2$, but ours increases faster and quickly becomes the best when $r > 2$. We also introduce three statistical measures of error ratios, *i.e.*, the mean and worst error ratio, and success rate, in Table. I. The proposed algorithm is superior to the competing algorithms in all three measures. Apart from quantitative assessments, Fig. 9 shows some qualitative examples of challenging cases. The proposed algorithm achieves more robust results on these cases, compared with the competing algorithms.

### B. Real Motion Blurred Images

Besides artificial blurred database, we further apply the proposed algorithm on real motion blurred images. For real images, the blurred cases are usually more complicated, including different depth of field (Fig. 10), mixture of human and scenes (Fig. 11), more complex and severer motions (Fig. 12, 13), *etc*. We compare our algorithm with recent methods, of which the implementations are available, *i.e.*, Krishnan *et al.* [42], Levin *et al.* [41], Michaeli & Irani [44] and Pan *et al.* [6].

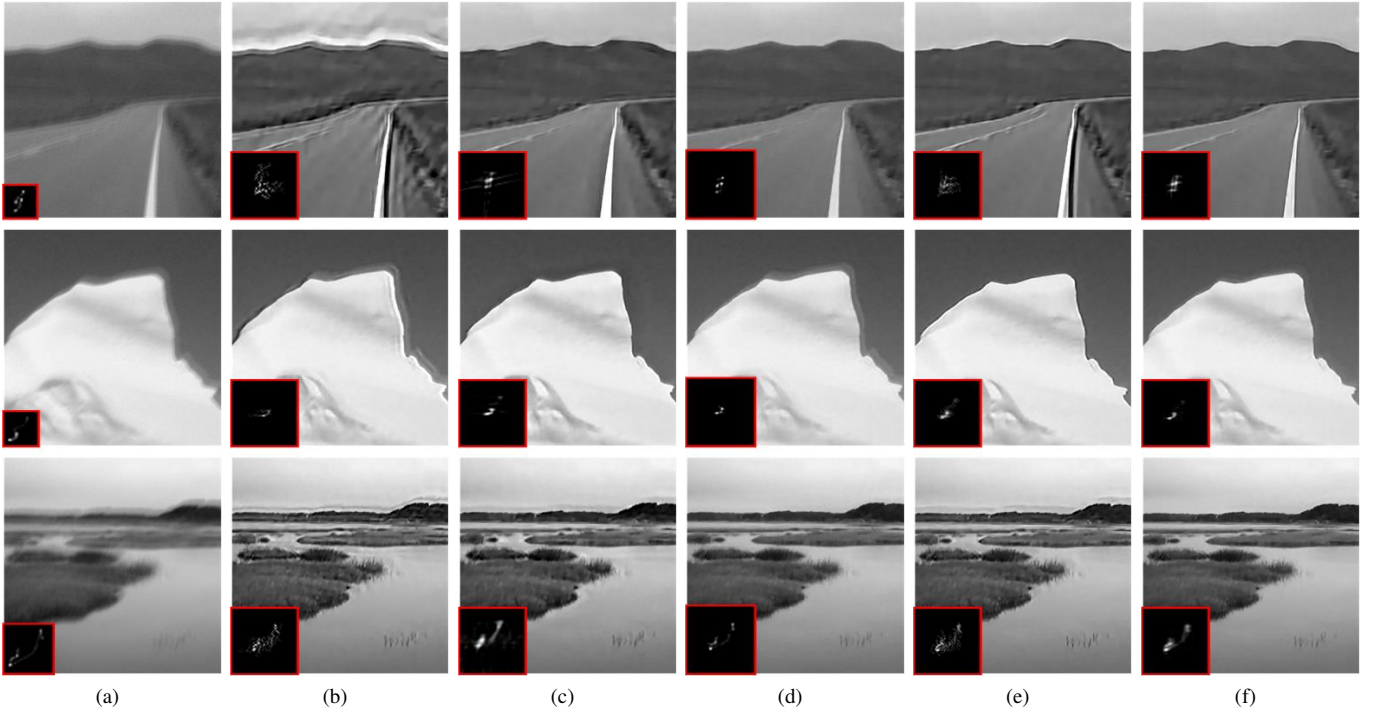

(a)          (b)          (c)          (d)          (e)          (f)

Fig. 9. *Three Deblurring Examples from Artificial Database*. (a) Blurry Image. (b) Sun *et al.* [26]. (c) Michaeli & Irani [44]. (d) Lai *et al.* [45]. (e) Pan *et al.* [6]. (f) The proposed algorithm. The images are better viewed in full size on computer screen.

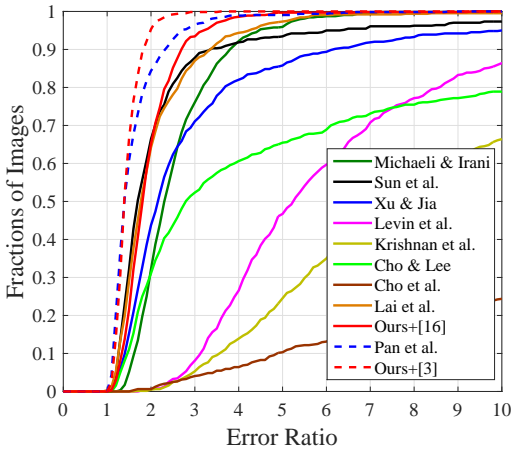

Fig. 8. Cumulative distribution of error ratios on artificial blurry database. Each algorithm estimates blur kernels and uses [18] (solid line) or [4] (dash line) as the final step non-blind deblurring.

In each experiment, all the algorithms are applied to estimate the blur kernel and then we use sophisticated non-blind image deblurring algorithm in [4] to restore latent sharp images. The proposed algorithm can robustly estimate blur kernels and results in fewer artifacts in the restored images. The visual comparisons show that our algorithm is obviously better than Krishnan *et al.* [42], Levin *et al.* [41] and Michaeli & Irani [44], and is competitive against Pan *et al.* [6] in experiments, as shown in Fig. 10–13.

### C. Gaussian Blurred Images

In Sec. V-C, we propose a specific acceleration for blind Gaussian blur deblurring in Algorithm 3. The experiments are set up by considering both the manual Gaussian blur and the unknown blur introduced by image scaling. In Fig. 14, the ground-truth image is blurred by Gaussian blur with $\sigma_b = 1.85$. In Fig. 15, the unknown blur is introduced before $2\times$ bilinear downsampling using "imresize" Matlab function. We apply Michaeli & Irani [44], Pan *et al.* [6], Algorithm 1 and Algorithm 3 to estimate the blur kernels. Afterwards, we use the classic hyper-Laplacian based non-blind deblurring algorithm [17] to restore the latent sharp image. The restored results, PSNRs and running time are reported in Fig. 14–15 and Table. II, respectively.

Algorithm 3 can estimate a blur kernel that is similar to the ground-truth kernel, and the restored images are obviously sharpened without artifacts, as shown in Fig. 14–15. Further, for the same or slightly better PSNR performance, the accelerated Algorithm 3 is significantly faster than the competing algorithms in Table. II. We note that the kernel restored by Algorithm 1 is the most similar to the ground-truth kernel. Overall, both the proposed algorithms outperform the competing methods in the specific Gaussian deblurring experiments.

### VII. CONCLUSION

The proposed RGTV is an effective prior to promote image sharpness and penalize blurriness. In this paper, we analyze its advantages in both nodal and graph frequency domain. In the nodal domain, RGTV can promote the bi-modal weight distribution of sharp images from blurry observations. In the graph frequency domain, RGTV enjoys desirable spectral properties, including robustness to noise and blur, strong PWS-preserving filtering and sharpness promotion. We design a robust blind image deblurring algorithm using RGTV. Experimental results demonstrate that the proposed algorithm can deal with various blind image deblurring scenarios, and the reconstructed sharp results are visually and numerically better than the state-of-the-art methods.

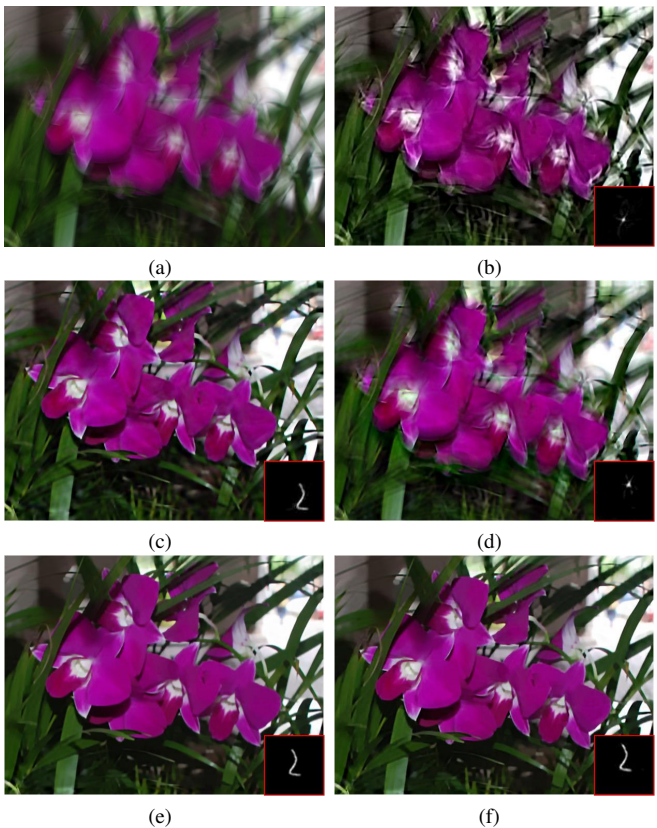

Fig. 10. *Real Blind Motion Deblurring Example*. Image size: $618 \times 464$, kernel size: $69 \times 69$. (a) Blurry image. (b) Krishnan *et al.* [42]. (c) Levin *et al.* [41]. (d) Michaeli & Irani [44]. (e) Pan *et al.* [6]. (f) The proposed Algorithm 1. The images are better viewed in full size on computer screen.

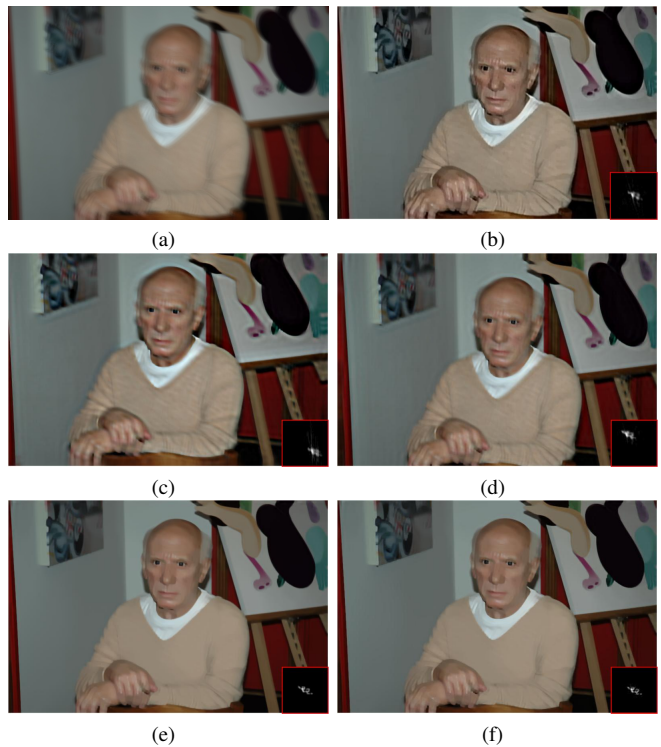

Fig. 11. *Real Blind Motion Deblurring Example*. Image size: $800 \times 532$, kernel size: $69 \times 69$. (a) Blurry image. (b) Krishnan *et al.* [42]. (c) Levin *et al.* [41]. (d) Michaeli & Irani [44]. (e) Pan *et al.* [6]. (f) The proposed Algorithm 1. The images are better viewed in full size on computer screen.

TABLE II
Quantitative performance of algorithm 3 compared with algorithm 1 on specific Gaussian-like blurry images.

| Examples | Images | PSNR(dB) | Time(s) |
|---|---|---|---|
| Fig. 14 | Blurred | 26.7 | - |
| | Micheali & Irani [44] | 30.6 | 1150.0 |
| | Pan *et al.* [6] | 30.4 | 421.3 |
| | Alg. 1 | 30.8 | 30.0 |
| | Alg. 3 | 30.8 | 7.8 |
| Fig. 15 | Up-sampled | 32.1 | - |
| | Micheali & Irani [44] | 32.6 | 1142.0 |
| | Pan *et al.* [6] | 33.0 | 428.3 |
| | Alg. 1 | 33.9 | 30.0 |
| | Alg. 3 | 33.8 | 7.9 |

The proposed algorithm with RGTV mainly focuses on uniform blur, *i.e.*, blur follows the convolution model (1) in the data fidelity term of (17). Leveraging our theoretical analysis on RGTV in both nodal and graph spectral domains, one can easily employ RGTV to solve non-uniform deblurring problems also, as long as there is a more sophisticated non-uniform blur model substituting for the current convolution model. In the future, we will extend our algorithm to solve non-uniform camera motion deblurring problems, by combining RGTV with the existing camera geometric model [46]–[48]. Moreover, as there are many fast-developing imaging technologies, such as light field imaging and multi-spectral imaging, we also would like to use RGTV for blind deblurring in these advanced imaging systems.

## Appendix A
### Derivation of weight function of GTV

As GTV is non-differentiable, we take the sub-derivative of GTV, resulting in the following sub-differential:

$$(\partial \|\mathbf{x}\|_{GTV})_{i,j} = \begin{cases} \frac{w_{i,j}}{|x_i - x_j|} \cdot (x_i - x_j), & x_i \neq x_j; \\ [-1, 1], & x_i = x_j. \end{cases} \quad (29)$$

Because the denominator of the weight term in the sub-differential (29), $|x_i - x_j|$, goes to 0 as $x_i$ approaches $x_j$, for numerical stability we approximate (29) with the following when $|x_i - x_j|$ is less than a small $\epsilon$; *i.e.*,

$$(\partial \|\mathbf{x}\|_{GTV}^{\epsilon})_{i,j} = \begin{cases} \frac{w_{i,j}}{|x_i - x_j|} \cdot (x_i - x_j), & |x_i - x_j| \geq \epsilon; \\ \frac{w_{i,j}}{\epsilon} \cdot (x_i - x_j), & |x_i - x_j| < \epsilon. \end{cases} \quad (30)$$

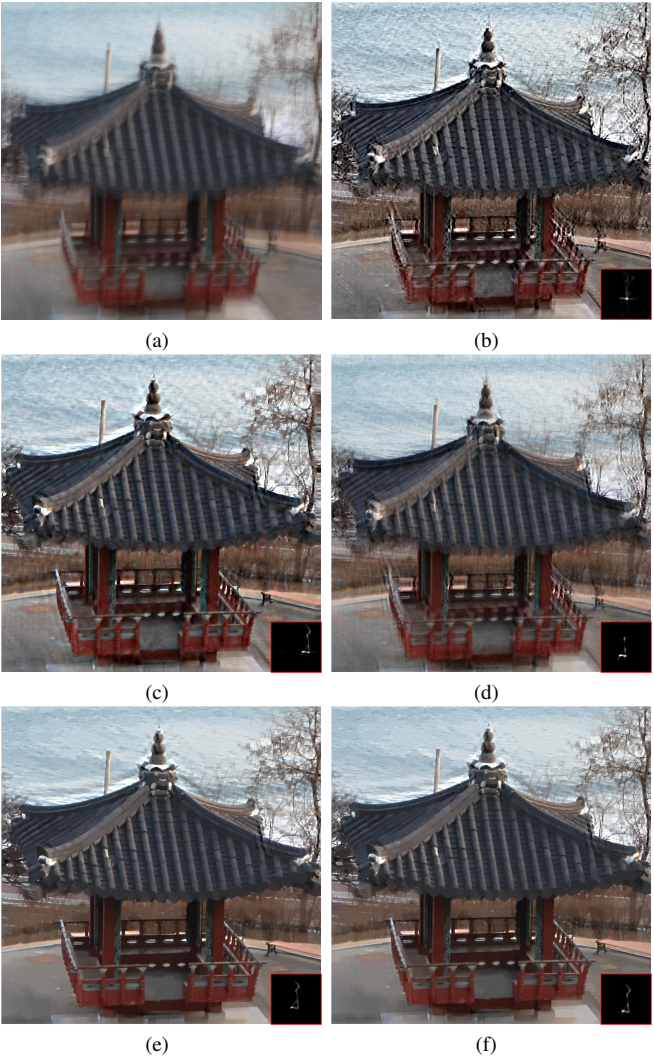

Fig. 12. *Real Blind Motion Deblurring Example*. Image size: $972 \times 966$, kernel size: $69 \times 69$. (a) Blurry image. (b) Krishnan *et al.* [42]. (c) Levin *et al.* [41]. (d) Michaeli & Irani [44]. (e) Pan *et al.* [6]. (f) The proposed Algorithm 1. The images are better viewed in full size on computer screen.

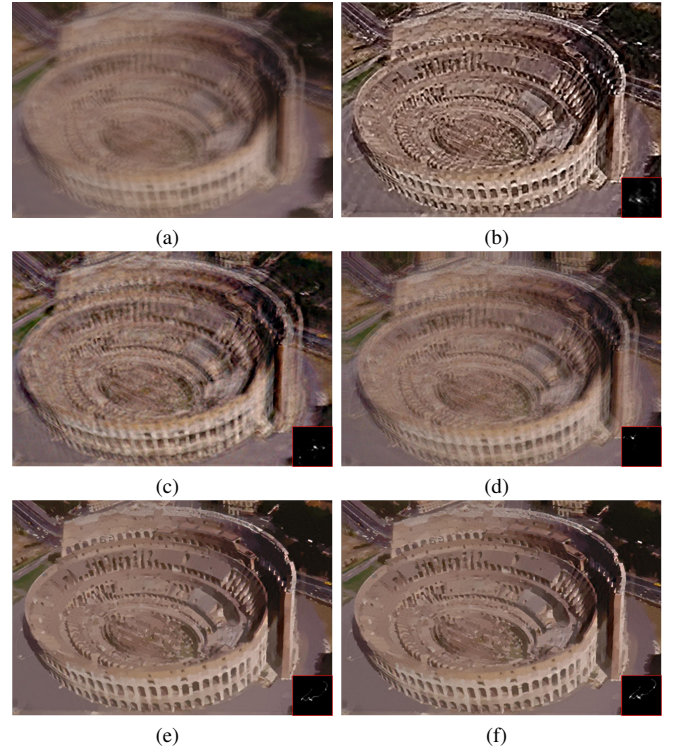

Fig. 13. *Real Blind Motion Deblurring Example*. Image size: $1229 \times 825$, kernel size: $95 \times 95$. (a) Blurry image. (b) Krishnan *et al.* [42]. (c) Levin *et al.* [41]. (d) Michaeli & Irani [44]. (e) Pan *et al.* [6]. (f) The proposed Algorithm 1. The images are better viewed in full size on computer screen.

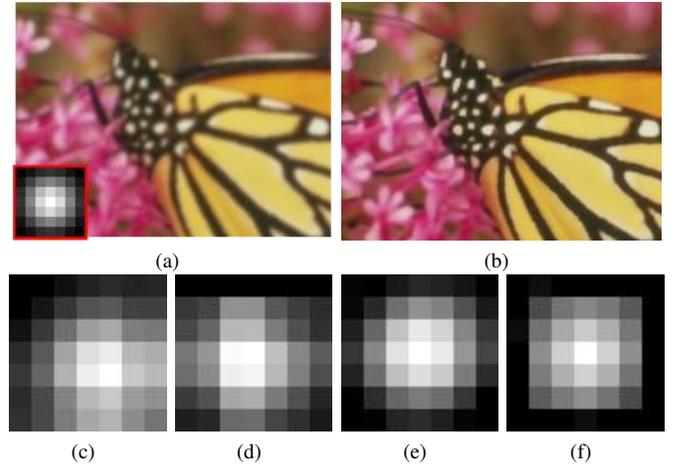

Fig. 14. *Unknown Gaussian Blur Example*. Image size: $768 \times 512$, kernel size: $7 \times 7$, $\sigma_b = 1.85$. (a) Blurry image and ground-truth kernel. (b) Restored image by the proposed Algorithm 3. (c) Restored kernel by Michaeli & Irani [44]. (d) Restored kernel by Pan *et al.* [6]. (e) Restored kernel by the proposed Algorithm 1. (f) Restored kernel by the proposed Algorithm 3.

where $\epsilon$ is a small constant close to 0. (30) is equivalent to earlier (11) and (12). (30) can actually be interpreted as the derivative of an upper-bound function of GTV:

$$(\|\mathbf{x}\|_{GTV}^{\epsilon})_{i,j} = \begin{cases} w_{i,j}|x_i - x_j|, & |x_i - x_j| \geq \epsilon; \\ \frac{w_{i,j}}{2\epsilon}(x_i - x_j)^2 + \frac{w_{i,j}\epsilon}{2}, & |x_i - x_j| < \epsilon. \end{cases} \quad (31)$$

where $(\|\mathbf{x}\|_{GTV}^{\epsilon})_{i,j}$ is the same as GTV $(\|\mathbf{x}\|_{GTV})_{i,j}$ except when $|x_i - x_j| < \epsilon$, when $(\|\mathbf{x}\|_{GTV}^{\epsilon})_{i,j}$ has a quadratic behavior with minimum at $\frac{w_{i,j}\epsilon}{2}$. The $l_1$ adjacency matrix $\mathbf{\Gamma}$ we use for GTV spectral analysis is based on this upper-bound weight function (31), which is reasonable, since (31) satisfies $\lim_{\epsilon \to 0} \|\mathbf{x}\|_{GTV}^{\epsilon} = \|\mathbf{x}\|_{GTV}$.

## APPENDIX B
## BOUNDEDNESS OF EIGENVALUES OF $\mathbf{L}_\Gamma$

*Lower bounded*: Because adjacency matrix $\mathbf{\Gamma}$ is a non-negative matrix and the definition of graph Laplacian matrix

$\mathbf{L}_\Gamma$ (13), $\mathbf{L}_\Gamma$ is a PSD matrix. Therefore, 0 is the lowest eigenvalue of $\mathbf{L}_\Gamma$ with $\mathbf{1}$ its eigenvector.

*Upper bounded*: We prove the upper boundedness based on Gershgorin Circle Theorem.

**Theorem 1** (Gershgorin Circle Theorem). *Let $\mathbf{A}$ be a $n \times n$ complex matrix with entries $a_{i,j}$. For $i \in \{1, ..., n\}$, $R_i$ is defined as $R_i = \sum_{j \neq i} |a_{i,j}|$. Every eigenvalue of $\mathbf{A}$ lies within*

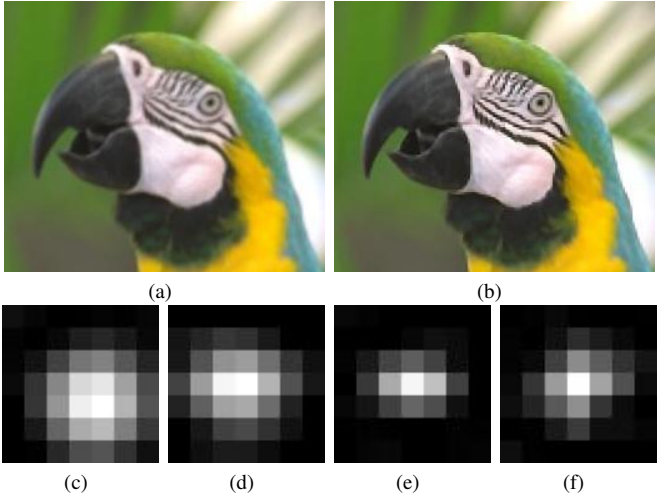

<div style="text-align:center">(a)      (b)</div>

<div style="text-align:center">(c)    (d)    (e)    (f)</div>

Fig. 15. *Unknown Blur Introduced Before Image Scaling.* Image size: $768 \times 512$. (a) $2\times$ up-sampled low-resolution image. (b) Restored image by the proposed Algorithm 3. (c) Restored kernel by Michaeli & Irani [44]. (d) Restored kernel by Pan *et al.* [6]. (e) Restored kernel by the proposed Algorithm 1. (f) Restored kernel by the proposed Algorithm 3.

at least one of the Gershgorin discs $D(a_{i,i}, R_i)$, *i.e.*,

$$|\lambda - a_{i,i}| \leq R_i \tag{32}$$

For matrix $\mathbf{A} = \mathbf{L}_\Gamma$, $a_{i,i} = \sum_{j \neq i} \gamma_{i,j} \leq \frac{d_i \max_j(w_{i,j})}{\epsilon} \leq \frac{d_i}{\epsilon}$ and $R_i = \sum_{j \neq i} |a_{i,j}| = \sum_{j \neq i} \gamma_{i,j} \leq \frac{d_i \max_j(w_{i,j})}{\epsilon} \leq \frac{d_i}{\epsilon}$. $d_i$ is the degree of node $i$. Therefore, we have

$$\max_i \lambda_i \leq \max_i (a_{i,i} + R_i) \leq \max_i (\frac{2d_i}{\epsilon}), \tag{33}$$

Since $d_i \leq n$, $\lambda$ is upper-bounded for fixed $\epsilon$.

## APPENDIX C
## PROOF OF POSITIVE DEFINITENESS

*Proof*: We prove $\mathbf{x}^T(\hat{\mathbf{K}}^T\hat{\mathbf{K}} + 2\beta \cdot \mathbf{L}_\Gamma)\mathbf{x} > 0$ $(\beta > 0)$, for $\forall \mathbf{x} \in \mathbb{R}^n$ and $\mathbf{x} \neq \mathbf{0}$. Here we set $\beta = 0.5$ without loss of generality.

$$\mathbf{x}^T(\hat{\mathbf{K}}^T\hat{\mathbf{K}} + \mathbf{L}_\Gamma)\mathbf{x} = \mathbf{x}^T(\hat{\mathbf{K}}^T\hat{\mathbf{K}})\mathbf{x} + \mathbf{x}^T\mathbf{L}_\Gamma\mathbf{x},$$
$$= \|\hat{\mathbf{K}}\mathbf{x}\|_2^2 + \sum_{i,j} \gamma_{i,j}(x_i - x_j)^2 \tag{34}$$

In (34), $\|\hat{\mathbf{K}}\mathbf{x}\|_2^2 \geq 0$. $\sum_{i,j} \gamma_{i,j}(x_i - x_j)^2 \geq 0$, for $\gamma_{i,j} \geq 0$ given the weight definition (12). Therefore, $\mathbf{x}^T(\hat{\mathbf{K}}^T\hat{\mathbf{K}} + \mathbf{L}_\Gamma)\mathbf{x}$ is a positive semi-definite matrix.

We further prove $\mathbf{x}^T(\hat{\mathbf{K}}^T\hat{\mathbf{K}} + \mathbf{L}_\Gamma)\mathbf{x}$ is strictly positive. We define two sets:

$$S_1 = \{\mathbf{x} | \mathbf{x}^T(\hat{\mathbf{K}}^T\hat{\mathbf{K}})\mathbf{x} = 0, \mathbf{x} \in \mathbb{R}^n, \mathbf{x} \neq 0\} \tag{35}$$

$$S_2 = \{\mathbf{x} | \mathbf{x}^T\mathbf{L}_\Gamma\mathbf{x} = 0, \mathbf{x} \in \mathbb{R}^n, \mathbf{x} \neq 0\} \tag{36}$$

As both $\hat{\mathbf{K}}^T\hat{\mathbf{K}}$ and $\mathbf{L}_\Gamma$ are already positive semi-definite, we only need to prove $S_1 \cap S_2 = \emptyset$.

$\mathbf{L}_\Gamma$ is a combinatorial graph Laplacian of a connected graph; by definition it must have one constant eigenvector corresponding to eigenvalue 0 and the multiplicity of eigenvalue 0 is 1. Therefore, $S_2 = \{\mathbf{1}\}$, where $\mathbf{1}$ is a constant vector with all elements 1.

For image blur caused by out-of-focus or motion, each element of blur kernel $k_i \geq 0$ and $\sum_i k_i = 1$. Therefore, $\hat{\mathbf{K}}$ is a non-negative matrix and $\hat{\mathbf{K}} \cdot \mathbf{1} = \mathbf{1}$, *i.e.*, $\lambda = 1$ is an eigenvalue of matrix $\hat{\mathbf{K}}$ and corresponding eigenvector is a constant vector $\mathbf{1}$. This property is intuitive; blurring a constant signal results in the signal itself under suitable boundary conditions. Therefore, $\mathbf{1} \notin S_1$ and $S_1 \cap S_2 = \emptyset$.

$S_1 \cap S_2 = \emptyset$ means that the quadratic forms of two positive semi-definite matrices, *i.e.*, $\mathbf{x}^T(\hat{\mathbf{K}}^T\hat{\mathbf{K}})\mathbf{x}$ and $\mathbf{x}^T\mathbf{L}_\Gamma\mathbf{x}$, cannot be 0 at the same time. Therefore, $\hat{\mathbf{K}}^T\hat{\mathbf{K}} + 2\beta \cdot \mathbf{L}_\Gamma (\beta > 0)$ is a positive definite matrix.